\documentclass[preprint,12pt, a4paper]{elsarticle}

\usepackage{amssymb}
\usepackage[utf8]{inputenc}
\usepackage{hyperref}
\usepackage{setspace}
\usepackage{multirow}
\usepackage{enumitem}

\hypersetup{colorlinks=false,pdfborder={0 0 0},
bookmarksopen,bookmarksdepth=3}

\journal{arXiv}

\begin{document}

\begin{frontmatter}



\title{\textit{Construe}: a software solution for the explanation-based 
interpretation of time series}


\author[esl]{T. Teijeiro}
\author[citius]{P. Félix}

\address[esl]{Embedded Systems Laboratory, École Polytechnique Fédérale de 
Lausanne (EPFL), Lausanne - 1015, Switzerland}
\address[citius]{Centro Singular de Investigación en Tecnoloxías Intelixentes 
(CITIUS), Universidade de Santiago de Compostela, Santiago de Compostela -
15782, Spain}

\begin{abstract}
This paper presents a software implementation of a general framework for time 
series interpretation based on abductive reasoning. The software provides a data 
model and a set of algorithms to make inference to the best explanation of a 
time series, resulting in a description in multiple abstraction levels of the 
processes underlying the time series. As a proof of concept, a comprehensive 
knowledge base for the electrocardiogram (ECG) domain is provided, so it can be 
used directly as a tool for ECG analysis. This tool has been successfully 
validated in several noteworthy problems, such as heartbeat classification or 
atrial fibrillation detection.
\end{abstract}

\begin{keyword}
Time Series Interpretation \sep Temporal Abstraction \sep Abductive Reasoning 
\sep ECG delineation \sep Arrhythmia Recognition \sep Python \sep WFDB
\end{keyword}

\end{frontmatter}




\section{Motivation and significance}
\label{sec:motivation}

%
%
%
%

Recently, a novel approach for time series interpretation has been 
proposed~\cite{Teijeiro18}, based on the principles of abductive reasoning and 
conceiving the interpretation task as a process of hypotheses formulation and 
testing in multiple abstraction levels. This proposal is essentially different 
from traditional approaches to pattern recognition, which in general pose time 
series interpretation as a sequence of classification problems in multiple 
levels, where results from lower levels are taken as the input for higher 
levels~\cite{Ganz15}. Standard classifiers behave as deductive systems, and 
once a decision has been made it cannot be changed or retracted within the same 
classifier. This causes that errors in the lower level classifiers are 
propagated upwards, resulting in a system that is weaker than any of 
the individual components.

An abductive approach, however, features a non-monotonic reasoning paradigm, 
which supports the amendment of any conclusion at any abstraction level in the 
search for the best global explanation of the observed evidence. The strategy is 
inspired by how humans identify and characterize the patterns appearing in a 
time series, leveraging both bottom-up and top-down reasoning to provide a joint 
result. As a consequence, abduction can guess the underlying processes from 
corrupted data or even in the temporary absence of data, by considering the 
context information from higher abstraction levels. This achieves greater 
robustness in the interpretation process and overcomes some of the most 
important weaknesses of traditional paradigms~\cite{Teijeiro18}.

These properties made it possible to successfully address two different 
problems in the domain of automatic electrocardiogram (ECG) analysis. On the one 
hand, the interpretations provided by \textit{Construe} are the basis of a new 
method for heartbeat classification~\cite{JBHI16} that significantly outperforms 
any other automatic approaches in the state-of-the-art, and even improves most 
of the assisted approaches that require expert aid. Also, a combination of 
\textit{Construe} with machine-learning algorithms obtained the highest score in 
the Physionet/Computing in Cardiology 2017 Challenge, outperforming the most 
popular techniques such as deep learning and random forests~\cite{Teijeiro18b}.

But our main motivation is to expand the use of this method and its adaptation 
to new problems and domains. In order to encourage this, we provide a free and 
open source reference implementation oriented to ECG applications. This 
implementation includes an abstract data model and a set of interpretation 
algorithms, as well as a knowledge base formalizing the standard clinical 
criteria for ECG analysis. This knowledge base supports building an 
interpretation of the ECG trace in the abstraction levels typically adopted in 
cardiology for explaining the physiological processes that take place in the 
heart muscle, including the electrical activation/recovery of the atria and the 
ventricles, and the different rhythm patterns that can be observed over time. 
Hence, this software can be directly used as an advanced ECG analysis tool, 
with a similar interface to that of other software packages such as 
ecg-kit~\cite{Demski16} or the WFDB Applications~\cite{Moody14}.

\section{Software description}
\label{sec:description}


\subsection{Software Functionalities}
\label{sec:functionalities}


In short, the \textit{Construe} project provides a domain-independent framework 
for the explanation-based interpretation of time series. This framework includes 
a mechanism for designing \textit{abstraction patterns}, and an interpretation 
algorithm that produces an explanation of a time series on the basis of the 
designed patterns.

\begin{figure}
\centering
\includegraphics[width=0.8\textwidth]{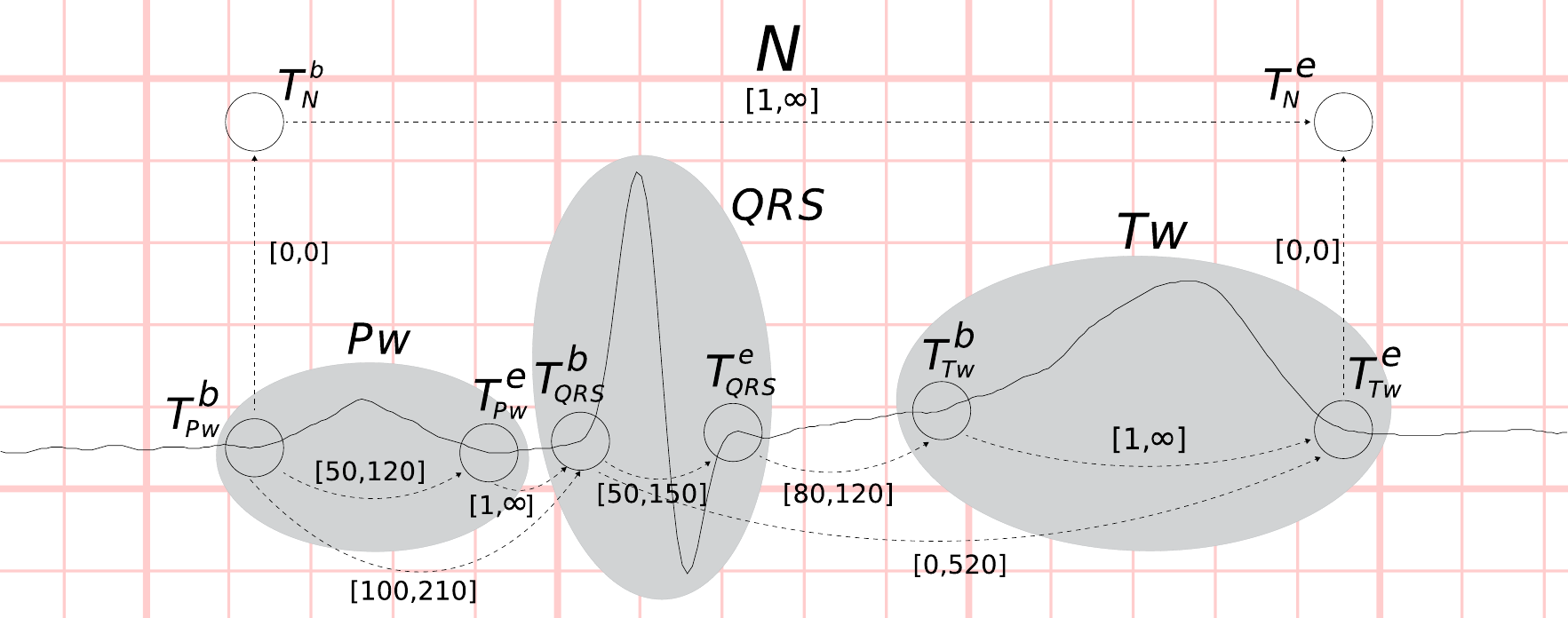}
\caption{Illustration of the \textit{Normal heart beat} abstraction pattern.}
\label{fig:normal_cycle}
\end{figure}

Let's begin with a simple illustration of these concepts. 
Figure~\ref{fig:normal_cycle} shows an example of an abstraction pattern that 
formalizes the notion of a normal heart beat as it can be observed in an ECG 
signal. Each heartbeat is observed as a sequence of three wave components: The P 
wave, which represents the electrical activation of the atria; the QRS complex, 
corresponding to the electrical activation of the ventricles; and the T wave, 
which is the signature of the electrical recovery of the ventricles. The 
pattern also defines some temporal constraints in the durations and 
separations between the endpoints of the waves.

To describe abstraction patterns, the framework relies on two main entities:

\begin{itemize}[itemsep=0pt, label=-]
 \item \textit{Observables} \texttt{<model.Observable>}: They describe the 
ontological knowledge of a particular domain in an object-oriented fashion, 
that is, as a hierarchy of classes. In the example above there are four 
observables: \textit{P wave}, \textit{QRS Complex}, \textit{T Wave} and 
\textit{Normal heart beat}.
 \item \textit{Abstraction grammars} \texttt{<model.PatternAutomata>}: They 
enable the description of potentially infinite sets of abstraction patterns, 
setting an explanatory relation between a hypothesis observable (in the example 
this would be the \textit{Normal heart beat}) and different temporal 
arrangements of evidence observables. Formally they are described as 
right-linear attributed grammars~\cite{Teijeiro18}, so they provide the same 
expressiveness as regular expressions. For simplicity, they are implemented 
as the equivalent finite automata, where non-terminals are replaced by states, 
and rules are described as transitions between states. These automata support a 
flexible definition of temporal and value constraints between the evidence and 
the hypothesis by means of user-defined functions.
\end{itemize}

For the definition of temporal constraints, the Simple Temporal Problem (STP) 
model has been adopted~\cite{Dechter91}. This model allows to constrain the 
distance between temporal variables by closed intervals, and supports a 
graph-based representation that can be efficiently analyzed in polynomial time. 
The knowledge representation model is sufficiently expressive to describe 
complex processes such as for example trigeminy arrhythmias~\cite{Teijeiro18}, 
and a tutorial on how to use the knowledge description tools is included in the 
project wiki page. 

Once a set of abstraction patterns for a specific problem has been defined, 
\textit{Construe} will receive as input a time series and will produce an 
interpretation. The interpretation algorithm is based on a heuristic search 
procedure inspired by the K-Best-First-Search algorithm~\cite{Edelkamp2011} that 
combines a set of reasoning modes to implement a hypothesize-and-test cycle 
guided by an attentional mechanism. As a result, the algorithm provides a set of 
annotated time intervals with the hypotheses explaining the input time series. 
\textit{Construe} can also work in online mode, processing the signal while it 
is being acquired.

On the other hand, in its present application as an end-user tool, this software 
includes a built-in knowledge-base that supports the interpretation of 
multi-lead ECG records in the MIT-BIH format~\cite{Moody14}. Given an ECG 
record, \textit{Construe} infers the set of hypotheses that best explain the 
observed signal. These resulting hypotheses describe the signal behavior in two 
abstraction levels: 1) conduction level, that provides a delineation for the P, 
QRS, and T waves; and 2) rhythm level, that gives the sequence of rhythm 
patterns, including normal sinus rhythm, bradycardia, tachycardia, extrasystole, 
couplet, rhythm block, bigeminy, trigeminy, atrial fibrillation, ventricular 
flutter and asystole. This result can be exported to a standard annotation file, 
enabling further analysis or visualization with external tools. The interface 
has been designed to be compatible with the well-known command-line tools from 
the WFDB library~\cite{Moody14}.

\subsection{Software Architecture}
\label{subsec:architecture}


\begin{figure}[t!!]
\centering
\includegraphics[width=0.8\textwidth]{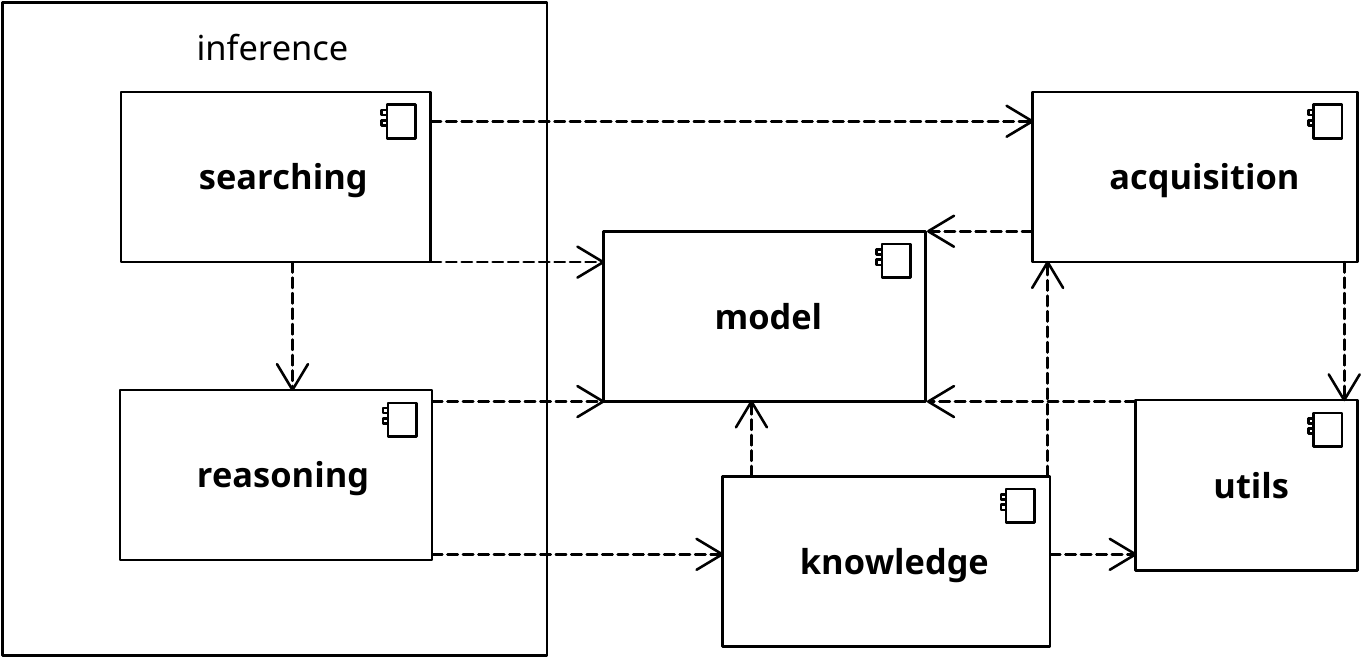}
\caption{Framework architecture}
\label{fig:arch}
\end{figure}

Figure~\ref{fig:arch} shows the main components of the framework, which roughly 
correspond to the top-level packages of the project. Each of these components 
are described below:
\begin{itemize}[itemsep=0pt, label=-]
 \item \textbf{inference:} Contains the implementation of the \textit{Construe} 
algorithm. It is composed by two modules: \textbf{searching}, that provides the 
heuristic search method; and \textbf{reasoning}, that implements the reasoning 
modes composing the hypothesize-and-test cycle. This component is 
domain-independent.
 \item \textbf{acquisition:} Provides global buffers for the acquisition of the 
time series to be interpreted, with support for online interpretation scenarios.
 \item \textbf{model:} Defines the general data model of the framework. This 
includes \textit{observables}, \textit{abstraction patterns}, 
\textit{abstraction grammars} and \textit{interpretations}. A model for the 
definition of STP networks~\cite{Dechter91} is also provided.
 \item \textbf{utils:} Miscellaneous utility modules, including signal 
processing, plotting and data-format manipulation routines.
 \item \textbf{knowledge:} This package provides all the domain-specific 
knowledge for ECG interpretation, namely the ontology of observables and the 
automata defining the abstraction patterns supporting all the conduction-level 
and rhythm-level hypotheses.
\end{itemize}

\section{Illustrative Examples}
\label{sec:examples}

%
%

\begin{figure}[h]
\centering
\includegraphics[width=\textwidth]{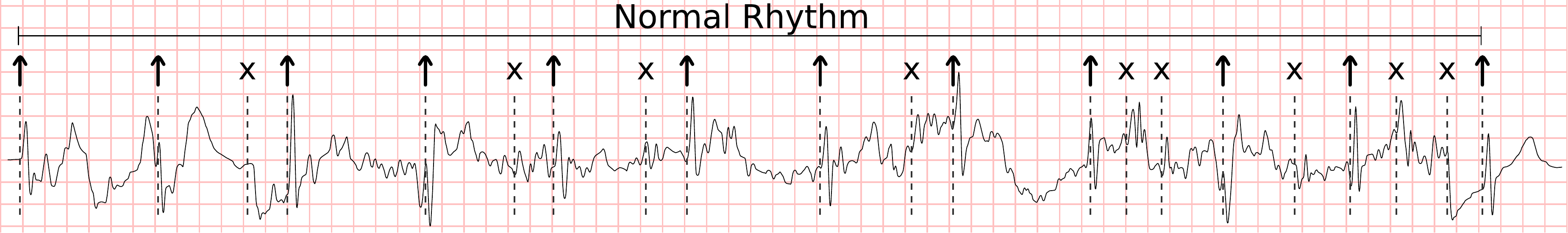}
\caption{Ignoring misleading evidence in a noisy ECG record. {\scriptsize 
[Source: Record A02080 from the PhysioNet Challenge 2017 dataset.]}
}
\label{fig:example}
\end{figure}

In the search for the best explanation of the initial evidence, 
\textit{Construe} is able both to ignore part of the evidence or to actively 
search for missing pieces, according to the known patterns. 
Figure~\ref{fig:example} shows how this ability to discard some evidence makes 
it possible to obtain useful interpretations from very distorted signals. The 
figure shows a noisy ECG segment, in which a standard heartbeat detector 
identifies a number of false positive beats (detections are the vertical lines). 
After the interpretation, \textit{Construe} concludes that the best explanation 
for the segment is that it corresponds to a normal rhythm, and that only the 
lines marked with arrows are actual heartbeats.

\begin{figure}[h]
\centering
\includegraphics[width=\textwidth]{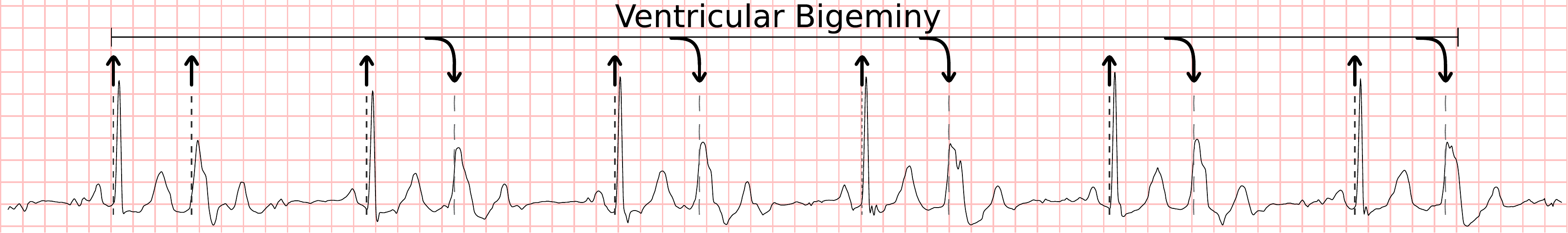}
\caption{Top-down discovery of missing beats in a ventricular bigeminy. 
{\scriptsize [Source: Record A01744 from the PhysioNet Challenge 2017 dataset.]}
}
\label{fig:example2}
\end{figure}

On the other hand, in Figure~\ref{fig:example2} we can see how the active 
search for missing information allows to find patterns that a common classifier 
would ignore. The figure shows a ventricular bigeminy pattern, in which every 
normal heartbeat is followed by an ectopic one, much more wider but with smaller 
amplitude. In this case, a standard annotator misses all but the first of the 
ectopic beats, so a normal classification would label the segment as an ectopic 
ventricular beat followed by a sinus bradycardia (low heart rate). However, in 
\textit{Construe} the first ectopic beat allows to hypothesize the presence of a 
possible bigeminy, and then actively look for the missing ectopic beats.

\section{Impact}
\label{sec:impact}

%
%
%
%
%

In the short term, the main impact of \textit{Construe} is expected to come from 
its function as an ECG interpretation tool, but not only for targeting the 
typical problems addressed by the automatic ECG processing community (continuous 
monitoring, outpatient follow-up, etc.). Besides the remarkable results we have 
already obtained in some noteworthy problems in this area~\cite{JBHI16, 
Teijeiro18b}, independent users have successfully used \textit{Construe} as a 
tool for ECG abstraction and feature extraction in large genome-wide studies to 
characterize the genetic variations that underlie cardiovascular diseases, aimed 
at a better understanding of the human physiology~\cite{Verweij18, VandeVegte18, 
Verweij19}. In these studies \textit{Construe} has been used to interpret tens 
of thousands of signals with very different properties (at rest, during 
exercise, with varying lead configurations, etc.), demonstrating the reliability 
of the algorithm beyond the common validation databases. And most importantly, 
in these works the interpretation of the ECG is not an end, but a means to reach 
new research hypotheses that go beyond an immediate individual diagnosis of the 
heart function.

The key factor behind the reliability of \textit{Construe} is the non-monotonic 
nature of the hypothesize-and-test cycle, making it possible to exploit the 
complementarity between bottom-up and top-down processing, in order to find the 
best explanation consistent with the evidence. As in perception, ECG 
interpretation is assumed to be mostly bottom up, although top-down processing 
has proven to be decisive to cope with noise, artifacts or ambiguities in the 
signal. Thus, in the longer term, we expect that the main impact of this 
software will be to spread the use of non-monotonic reasoning techniques for the 
abstraction of temporal information. This is especially suited for scenarios 
requiring a continuous interpretation of low quality sensory data, and that are 
particularly frequent in emerging Internet of Things (IoT) 
contexts~\cite{Alam17}.

\section{Conclusions}
\label{sec:conclusion}

The \textit{Construe} software aims to provide a framework for the development 
of knowledge-based solutions oriented to time series interpretation according to 
an abductive reasoning scheme. This framework defines a general data model for 
knowledge representation and a set of domain-independent interpretation 
algorithms, and it has been successfully tested on different problems in the ECG 
domain.

The main advantages of this approach are 1) the explainability of the models and 
of the interpretation results, a key advantage to gain the trust of human 
experts; and 2) the robustness to the presence of noise and artifacts, which 
makes it adequate for poorly controlled interpretation environments. 

As for the evolution of this project, besides deepening into the interpretation 
of biological signals to address more complex problems and to include new signal 
types and multi-parameter signals, efforts will be devoted to the integration of 
machine learning strategies for knowledge definition and adaptation, as well as 
to the improvement of the computational performance of the algorithms in 
special-purpose architectures. Also, we are confident that the generality of 
this framework will allow the scientific community to find new problems and 
domains of application.

\section*{Acknowledgements}
\label{sec:ack}

This work was supported by the Xunta de Galicia and the European Regional 
Development Fund (ERDF) under Grant No. 2016-2019-ED431G/08, and by the Human 
Brain Project (HBP) SGA2 (GA No. 785907).




\bibliographystyle{plain}
\bibliography{bibliography}

\begin{thebibliography}{10}

\bibitem{Alam17}
F.~Alam, R.~Mehmood, I.~Katib, N.~N. Albogami, and A.~Albeshri.
\newblock {Data Fusion and IoT for Smart Ubiquitous Environments: A Survey}.
\newblock {\em IEEE Access}, 5:9533--9554, 2017.

\bibitem{Dechter91}
M.~Dechter, J.~Meiri, and J.~Pearl.
\newblock {Temporal constraint networks}.
\newblock {\em Artificial Intelligence}, 49:61--95, 1991.

\bibitem{Demski16}
A.~Demski and M.~Llamedo.
\newblock {ecg-kit: a Matlab Toolbox for Cardiovascular Signal Processing}.
\newblock {\em Journal of Open Research Software}, 4(1), 2016.

\bibitem{Edelkamp2011}
S.~Edelkamp and S.~Schr{\"o}dl.
\newblock {\em {Heuristic Search: Theory and Applications}}.
\newblock Morgan Kaufmann, 2011.

\bibitem{Ganz15}
F.~Ganz, D.~Puschmann, P.~Barnaghi, and F.~Carrez.
\newblock {A Practical Evaluation of Information Processing and Abstraction
  Techniques for the Internet of Things}.
\newblock {\em IEEE Internet of Things Journal}, 2(4), 2015.

\bibitem{Moody14}
G.~B. Moody.
\newblock {WFDB Applications Guide, 10th edition}.
\newblock \url{http://www.physionet.org/physiotools/wag/}, 2014.
\newblock [Online; accessed 05/06/2019].

\bibitem{Teijeiro18}
T.~Teijeiro and P.~F{\'e}lix.
\newblock {On the adoption of abductive reasoning for time series
  interpretation}.
\newblock {\em Artificial Intelligence}, 262:163--188, 2018.

\bibitem{JBHI16}
T.~Teijeiro, P.~F{\'e}lix, J.~Presedo, and D.~Castro.
\newblock {Heartbeat classification using abstract features from the abductive
  interpretation of the ECG}.
\newblock {\em IEEE Journal of Biomedical and Health Informatics}, 22(2), 2018.

\bibitem{Teijeiro18b}
T.~Teijeiro, C.~A. Garc\'{\i}a, D.~Castro, and P.~Felix.
\newblock {Abductive reasoning as the basis to reproduce expert criteria in ECG
  atrial fibrillation identification}.
\newblock {\em Physiological Measurement}, 39(6), 2018.

\bibitem{VandeVegte18}
Yordi~J. van~de Vegte, Pim van~der Harst, and Niek Verweij.
\newblock {Heart Rate Recovery 10 Seconds After Cessation of Exercise Predicts
  Death}.
\newblock {\em Journal of the American Heart Association}, 7(8), 2018.

\bibitem{Verweij19}
N.~Verweij, J.W. Benjamins, M.P. Morley, Y.~van~de Vegte, A.~Teumer,
  T.~Trenkwalder, W.~Reinhard, T.~P. Cappola, and P.~van~der Harst.
\newblock {The genetic makeup of the electrocardiogram}.
\newblock {\em bioRxiv preprint}, 2019.

\bibitem{Verweij18}
N.~Verweij, Y.~J. {Van De Vegte}, and P.~{Van Der Harst}.
\newblock {Genetic study links components of the autonomous nervous system to
  heart-rate profile during exercise}.
\newblock {\em Nature Communications}, 9(1):898, 2018.

\end{thebibliography}

\break
\section*{Required Metadata}

\section*{Current code version}


\begin{table}[!!h]
\begin{tabular}{|l|p{6.5cm}|p{6.5cm}|}
\hline
\textbf{Nr.} & \textbf{Code metadata description} & \textbf{Please fill in this column} \\
\hline
C1 & Current code version & v2.1 \\
\hline
C2 & Permanent link to code/repository used for this code version & 
\url{https://github.com/citiususc/construe/archive/v2.1.zip} \\
\hline
C3 & Legal Code License   & AGPL v3 \\
\hline
C4 & Code versioning system used & git \\
\hline
C5 & Software code languages, tools, and services used & Python 3 \\
\hline
C6 & Compilation requirements, operating environments \& dependencies & - Python 
3 installation with the following packages:
\begin{itemize} 
\setlength{\itemsep}{1pt}
\setlength{\parskip}{1pt}
\setlength{\parsep}{1pt}
 \item \href{https://pypi.python.org/pypi/sortedcontainers}{sortedcontainers}
 \item \href{https://pypi.python.org/pypi/numpy}{numpy}
 \item \href{https://pypi.python.org/pypi/python-dateutil}{python-dateutil}
 \item \href{https://pypi.python.org/pypi/scipy}{scipy}
 \item \href{https://pypi.python.org/pypi/scikit-learn/0.18.1}{scikit-learn 
v0.18.1}
 \item \href{https://pypi.python.org/pypi/PyWavelets}{PyWavelets}
 \item \href{https://pypi.python.org/pypi/matplotlib}{matplotlib}
 \item \href{https://pypi.python.org/pypi/networkx}{networkx}
 \item \href{https://pypi.python.org/pypi/pygraphviz}{pygraphviz}
\end{itemize}
- \href{http://www.physionet.org/physiotools/wfdb.shtml}{WFDB Software 
package}

- \href{https://www.graphviz.org/}{Graphviz}
\\
\hline
C7 & If available Link to developer documentation/manual & 
\url{https://github.com/citiususc/construe} \\
\hline
C8 & Support email for questions & 
\href{mailto:tomas.teijeiro@epfl.ch}{tomas.teijeiro@epfl.ch}\\
\hline
\end{tabular}
\caption{Code metadata.}
\label{tab:code} 
\end{table}

\break
\section*{Current executable software version}


\begin{table}[!!h]
\begin{tabular}{|l|p{6.5cm}|p{6.5cm}|}
\hline
\textbf{Nr.} & \textbf{(Executable) software metadata description} & \textbf{Please fill in this column} \\
\hline
S1 & Current software version & v2.1 \\
\hline
S2 & Permanent link to executables of this version  & 
\url{https://github.com/citiususc/construe/archive/v2.1.zip}\\
\hline
S3 & Legal Software License & AGPL v3 \\
\hline
S4 & Computing platforms/Operating Systems & GNU/Linux \\
\hline
S5 & Installation requirements \& dependencies & - Python 3 installation with 
the following packages:
\begin{itemize} 
\setlength{\itemsep}{1pt}
\setlength{\parskip}{1pt}
\setlength{\parsep}{1pt}
 \item \href{https://pypi.python.org/pypi/sortedcontainers}{sortedcontainers}
 \item \href{https://pypi.python.org/pypi/numpy}{numpy}
 \item \href{https://pypi.python.org/pypi/python-dateutil}{python-dateutil}
 \item \href{https://pypi.python.org/pypi/scipy}{scipy}
 \item \href{https://pypi.python.org/pypi/scikit-learn/0.18.1}{scikit-learn 
v0.18.1}
 \item \href{https://pypi.python.org/pypi/PyWavelets}{PyWavelets}
 \item \href{https://pypi.python.org/pypi/matplotlib}{matplotlib}
 \item \href{https://pypi.python.org/pypi/networkx}{networkx}
 \item \href{https://pypi.python.org/pypi/pygraphviz}{pygraphviz}
\end{itemize}
- \href{http://www.physionet.org/physiotools/wfdb.shtml}{WFDB Software 
package}

- \href{https://www.graphviz.org/}{Graphviz}
\\
\hline
S6 & Link to user manual & \url{https://github.com/citiususc/construe} \\
\hline
S7 & Support email for questions & 
\href{mailto:tomas.teijeiro@epfl.ch}{tomas.teijeiro@epfl.ch}\\
\hline
\end{tabular}
\caption{Software metadata.}
\label{tab:software} 
\end{table}

\end{document}